\documentclass[a4paper]{article}

\usepackage[T1]{fontenc}

\usepackage{graphicx}
\usepackage{amsmath,amsfonts,amssymb}

\usepackage{bbm}

\usepackage{algorithm}
\usepackage{algorithmic}

\usepackage[caption=false]{subfig}

\usepackage{multirow}

\usepackage{authblk}

\usepackage[hidelinks]{hyperref}

\usepackage{orcidlink}

\begin{document}
\title{\textbf{Improving Insurance Catastrophic Data with Resampling and GAN Methods}}
%

%
\author{Norbert Dzadz}
\affil{Faculty of Mathematics and Information Science, Warsaw University of Technology,\\ Koszykowa 75, 00-662 Warsaw, Poland\\
\emph{email:} \href{mailto:norbert.dzadz.stud@pw.edu.pl}{norbert.dzadz.stud@pw.edu.pl}}

\author{Maciej Romaniuk\orcidlink{0000-0001-9649-396X}\footnote{Corresponding author}}

\affil{Systems Research Institute PAS, Newelska 6, 01-447 Warszawa, Poland\\
\emph{email:} \href{mailto:mroman@ibspan.waw.pl}{mroman@ibspan.waw.pl}
}
\affil{WIT Academy,  Newelska 6, 01-447 Warszawa, Poland}

\maketitle              
\begin{abstract}
The precise and large dataset concerning catastrophic events is very important for insurers.
To improve the quality of such data three methods based on the bootstrap, bootknife, and GAN algorithms are proposed.
Using numerical experiments and real-life data, simulated outputs for these approaches are compared based on the mean squared (MSE) and mean absolute errors (MAE).
Then, a direct algorithm to construct a fuzzy expert's opinion concerning such outputs is also considered. 

\end{abstract}
\section{Introduction}
\label{intro}

The quality of the dataset consisting of values and moments of catastrophic events (like earthquakes or floods) is very important for the insurer.
These data allow for developing, issuing, and pricing insurance instruments like policies, reinsurance contracts, catastrophe bonds, etc., which are then embedded into the insurer's portfolio \cite{Romaniuk20203}.
Usually, such data do not have ``proper'' quality -- the catastrophic events are rather rare (compared with the ``standard'' car accidents), and there are significant problems with modeling of the claim values because of extreme events, the necessity of using heavy-tailed distributions, etc. \cite{GIURICICH2019498,nhess-12-535-2012}.

The main aim of this article is to propose and compare three methods that could multiply (in some sense) the quantity and improve the quality of the catastrophic data, namely the bootstrap, bootknife, and GAN approaches.
The first two methods are rather classical ones \cite{Efron}, but the last one is considered as  state-of-the-art for developing generative models and is widely used in many real-life areas \cite{NIPS2014_5ca3e9b1}.
These methods are compared on the basis of the fitting model trained on the learning set and then evaluated on the test set.
As the goal, we use the value of the claim process \cite{doi:10.1142/7431}.
Apart from its ``crisp'' (i.e. real-valued) estimator, we propose its imprecise counterpart based on the left-right fuzzy numbers \cite{ban_coroianu_pg}.
An  ``expert's opinion'' obtained in this way incorporates additional information that can be valuable for the insurer.

This paper is organized as follows.
In Sect. \ref{pre}, the preliminaries concerning notation and the applied simulation methods are recalled.
The numerical experiments are described in Sect. \ref{nusi}, together with the comparison of the results and proposition of formulation of a fuzzy expert's opinion.
Some final remarks and possible future research are summarized in Sect. \ref{con}.

\section{Preliminaries}
\label{pre}

In this section, we present the necessary notation and introduction to the methods discussed further in the paper.

\subsection{Risk process}
\label{riskpro}

The \textbf{classical risk reserve process} $R_t$ (see, e.g., \cite{doi:10.1142/7431}) is defined as a model of the financial reserves of an insurer depending on time $t$, i.e.
\begin{equation}\label{rom:1}
	R_t=u+pt-S_t
\end{equation}
where $u$ is an initial reserve of the insurer, $p$ is a rate of premiums paid by the insureds per unit of time and $S_t$ is a claim process of the form
\begin{equation}\label{rom:2}
	S_t=\sum_{i=1}^{N_t} C_i
\end{equation}
where $C_1,C_2, \ldots$ are \emph{iid} random values of the claims, and $N_t$ is a random number of the claims till time $t$.
These claims are traditionally modeled by some probability distributions. 
We focus on the exponential distribution with the parameter $\lambda$ (abbreviated further as Exp $(\lambda) $), lognormal -- with the parameters $\mu, \sigma^2$ (LN $(\mu, \sigma^2 )$), Gamma -- with the shape $\alpha$ and scale $\beta$ parameters ($\Gamma (\alpha, \beta)$), and Weibull -- with the shape $k$ and scale $\lambda$ parameters (Weib $(k, \lambda)$).
These probability distributions are widely used in the insurance literature, e.g., \cite{Chernobai2006,Romaniuk201733} and in other real-life applications, e.g., \cite{DEMCart}.

The process $N_t$ is usually driven by the homogeneous or the nonhomogeneous Poisson process (abbreviated further on as HPP or NHPP, respectively), e.g., \cite{GIURICICH2019498,Romaniuk201733,Romaniuk2019111}.
In the following, we apply the NHPP with the sinusoidal \cite{Chernobai2006}
\begin{equation}
    \lambda (t) = \lambda_0 + \lambda_1 2 \pi \sin \left ( 2 \pi (t - \lambda_2) \right )
\end{equation}
and Power Law \cite{PowerLaw}
\begin{equation}
    \lambda (t) = \lambda_0 \lambda_1 t^{\lambda_1 - 1}
\end{equation}
intensity functions.

Apart from the behavior of  $S_t$, the insurer is also interested in an evaluation of \textbf{probability of an ultimate ruin} (i.e. a ruin with an infinite time horizon), which is given as
\begin{equation}
    \psi (u) = \Pr \left( \inf_{t \geq 0} R_t <0 \right )
\end{equation}
and \textbf{probability of a ruin} before the time $T$ (i.e. a ruin with a finite time horizon)
\begin{equation}
\label{probruinfintime}
    \psi (u,T) = \Pr \left( \inf_{t \in [0, T]} R_t <0 \right ) .
\end{equation}

\subsection{GAN and resampling methods}
\label{gainandres}

The \textbf{Generative Adversarial Network} (GAN, \cite{NIPS2014_5ca3e9b1}) consists of two neural networks: a generator and discriminator, that compete with each other in a zero-sum game.
After training on an initial set, this method generates new data that ``mimics'' the statistical properties of the training set \cite{GANart}.
This approach is considered state-of-the-art for developing generative models and is widely used for various kinds of data, including pictures and tabular data.
For this second case, some extensions of the original method were proposed, including incorporation of the expert knowledge as in the \textbf{DATGAN} method \cite{lederrey2022datgan}. 

The \textbf{bootstrap} proposed by Efron \cite{Efron} is a widely used resampling method.
Based on the initial sample $\mathbb{X} = (X_1, \ldots, X_n)$ with $n$ values, it generates $B$ secondary (bootstrap) samples $\mathbb{X}^{*j} = (X^{*j}_1, \ldots, X^{*j}_n)$, where $j=1, \ldots, B$, consisting of values randomly drawn from $\mathbb{X}$ with repetitions.
These bootstrap samples can be then applied, e.g., to better estimate the standard error of some estimator, or to approximate the p-value of a statistical test, also in the case of fuzzy numbers \cite{Grzegorzewski20201650,Grzegorzewski2022285,RJ-2023-036}.
There exist many extensions of the classical bootstrap, including the smoothed bootstrap and \textbf{bootknife} \cite{Hesterberg2002UnbiasingTB}.
In this second approach, the values for each bootstrap sample are randomly drawn as in the bootstrap, but only from $n-1$ possible values of the initial sample.
Firstly, for each bootstrap sample the value $j \in \{1, \ldots , n \}$ is randomly selected, and then this sample is generated based only on $\mathbb{X}^{-j} = (X_1, \ldots, X_{j-1}, X_j, \ldots, X_n)$.

\subsection{Catastrophic data}
\label{catada}

In our analysis, we use the data covering catastrophic events and the related claims for the years 1990--2022 in  North America from the EM-DAT dataset (\url{https://www.emdat.be/}) maintained by the Centre for Research on the Epidemiology of Disasters (CRED).
We will be interested in the pairs $(\mathbb{C},\mathbf{t}) = \{ (C_i, t_i)_{i=1} \}$, where $C_i$ stands for the value of the $i$-th claim in {\$}1000 adjusted to the inflation rate, and $t_i$ denotes time of its occurrence.
Additionally, only material claims (i.e. the values of the claims greater than zero) are considered further on, and 5{\%} of the highest claims (i.e., the easily spotted outliers) are removed from our dataset.
The data are divided into the training (the years 1990--2012, 375 cases) and test (the years 2013--2022, 138 cases) sets.

\subsection{Fuzzy numbers}

We provide only some necessary concepts and notations concerning fuzzy numbers.
For a more detailed introduction, see, e.g., \cite{ban_coroianu_pg}.

A \textbf{fuzzy number} (abbreviated further as FN) is an imprecise value characterized by a mapping $\tilde{A}:\mathbb{R}\to [0,1]$ (a \textbf{membership function}), such that its $\alpha$-cut defined by
\begin{equation}
\tilde{A}_{\alpha}=\begin{cases}
\{x\in\mathbb{R}:\tilde{A}(x)\geqslant\alpha\} & \text{if}\quad \alpha\in (0,1], \\
cl\{x\in\mathbb{R}:\tilde{A}(x)>0\} & \text{if}\quad \alpha=0,
\end{cases} \label{eq_acut}
\end{equation}
is a nonempty compact interval for each $\alpha\in [0,1]$. Operator $cl$ in \eqref{eq_acut} denotes the closure. Every FN is completely characterized both by its membership function $\tilde{A}(x)$ and a family of $\alpha$-cuts $\{\tilde{A}_{\alpha}\}_{\alpha\in [0,1]}$. There are two special $\alpha$-cuts: the \textbf{core} $\tilde{A}_1=\mathrm{core}(\tilde{A})$, which contains all values fully compatible with the concept described by $\tilde{A}$, and  the \textbf{support} $\tilde{A}_0=\mathrm{supp}(\tilde{A})$  for which values are compatible to some extent with the concept modeled by $\tilde{A}$. A family of all FNs will be denoted by $\mathbb{F}(\mathbb{R})$.

There exist many kinds of membership functions, e.g., the \textbf{LR-fuzzy numbers} (denoted further on as LRFNs) are defined by
\begin{equation} 
\tilde{A}(x)=
\begin{cases}
 0 & \text{if}\quad x < a_1,  \\
 L \left( \frac{x-a_1}{a_2 - a_1}\right) & \text{if}\quad a_1 \leqslant x < a_2 ,  \\
 1 & \text{if}\quad a_2 \leqslant x < a_3 , \\
 R \left( \frac{a_4 - x}{a_4 - a_3}\right) & \text{if}\quad a_3 \leqslant x < a_4 , \\
 0 & \text{if}\quad x \geq a_4,  
\end{cases}
\label{eq:LFfn}
\end{equation} 
where $L, R: [0,1] \rightarrow [0,1]$ are continuous and strictly increasing functions such that $L(0)=R(0)=0$ and $L(1)=R(1)=1$, and $a_1,a_2,a_3,a_4\in\mathbb{R}$, where $a_1\leqslant a_2\leqslant a_3\leqslant a_4$. 

\section{Numerical experiments}
\label{nusi}

In this section, different methods to improve the fitting of the models based on the catastrophic data (see Sect. \ref{catada}) are compared.
Then, we discuss a fuzzy approach to construct an expert's opinion concerning the value of the claim process.

\subsection{Comparison of the methods}
\label{noresap}

We start by using the dataset without its additional resampling (abbreviated further on as \emph{no-resampling}).
Then, the maximum likelihood estimators (MLEs) were calculated for the probability distributions of the claims values $C_i$ (see Sect.~\ref{riskpro}) based on the training dataset.
The results can be found in the first column of Table \ref{tab_values_claims}.
According to the p-values of the Kolmogorov-Smirnov and Cramer von Mises tests, only the LN distribution could not be rejected as the distribution modeling $C_i$.
However, to strengthen our reasoning the models and their errors for all of the considered distributions will be compared in the following.

\begin{table}[htb]
\centering
\caption{Comparison of the estimated parameters for the values of the claims.}
\renewcommand{\arraystretch}{1.1}
\begin{tabular}{|l|c|c|c|c|}
\hline
Distribution       & No-resampling & Bootstrap & Bootknife & GAN \\
\hline
Exp $(\lambda) $ & $  1.667 \cdot 10^{-6} $ & $  1.68 \cdot 10^{-6} $  & $  1.668 \cdot 10^{-6} $  & $  1.587 \cdot 10^{-6} $  \\
LN $(\mu, \sigma^2 )$ & $12.47, 2.07$  &  $12.46, 2.08$ & $12.47, 2.06$  &  $12.68, 1.06$ \\
$\Gamma (\alpha, \beta)$ & $0.72, 1.2 \cdot 10^{-6}$  & $0.72, 1.22 \cdot 10^{-6}$  & $0.72, 1.21 \cdot 10^{-6}$  & $0.88, 1.39 \cdot 10^{-6}$  \\
Weib $(k, \lambda)$ & $0.78, 514170.56$  & $0.79, 513403.98$  &  $0.79, 514926.24$ &  $0.88, 587007.16$ \\
 \hline
\end{tabular}
\renewcommand{\arraystretch}{1}
\label{tab_values_claims}
\end{table}

To obtain the estimators for the intensity function $\lambda (t)$ of the NHPP process for the number of claims $N_t$, the problem
\begin{equation}
\label{nhppestim}
    \min_{\lambda^*} \left( N_t - \Lambda (t) \right)
\end{equation}
in regard with the parameters $\lambda^*$ of the cumulated intensity function $\Lambda (t)$ for all $t \in (0,T)$ was solved with the Levenberg-Marquardt algorithm \cite{5451114}.
The results can be found in Table \ref{tab_int_claims}.

\begin{table}[htb]
\centering
\caption{Comparison of the estimated parameters for the intensity of the claims.}
\begin{tabular}{|l|c|c|}
\hline
Intensity function       & No-resampling/Bootstrap/Bootknife & GAN \\
\hline
Sinusoidal $(\lambda_0,\lambda_1, \lambda_2)$ &  $16.668, 11.076, 1.004$ & $16.61, 10.32, 0.98$  \\
Power Law $(\lambda_0,\lambda_1)$ & $32.86, 0.772$ & $31.93, 0.78 $ \\
 \hline
\end{tabular}
\label{tab_int_claims}
\end{table}

To compare the errors of the considered models on the test set, two widely used measures were applied, namely the Mean Squared Error (MSE) and Mean Absolute Error (MAE).
They are given by
\begin{equation}
    \text{MSE} = \frac{1}{n} \sum_{i=1}^{n} \left( S_T - \hat{S}_{T,i} \right)^2 , \text{MAE} = \frac{1}{n} \sum_{i=1}^{n} \left | S_T - \hat{S}_{T,i} \right | ,
\end{equation}
where $n$ is the number of the generated trajectories, $S_T$ is the true value of the claim process till time $T$, and $\hat{S}_{T,i}$ is the simulated value of the claim process for the $i$-th trajectory.
In the following experiments, $n=1000$ is set to decrease the randomness of the obtained results.

The estimated errors for different pairs, which consist of the probability distribution of $C_i$ and intensity function $\lambda (t)$, of the models can be found in Table~\ref{tab_errors}.
For the MSE, the respective values are given in {\$}$10^{16}$, and for the MAE -- in {\$}$10^{8}$.
It seems that the Power Law function gives a better fit, especially with the Weibull distribution.
When the probability of a ruin  \eqref{probruinfintime} is estimated for this best model, it is equal to 0.001 for 5 years, and 0.002 for 10 years horizons, when the initial reserve $u$ is set to {\$}$10^9$, and $p$ is equal to 0.3 (see also \cite{Chernobai2006} for the similar approach).

\begin{table}[htb]
\centering
\caption{Comparison of the errors of the models.}
\begin{footnotesize}

\begin{tabular}{|l|cc|cc|cc|cc|}
\hline
\multirow{2}{*}{Model}& \multicolumn{2}{|c|}{No-resampling}  & \multicolumn{2}{|c|}{Bootstrap} & \multicolumn{2}{|c|}{Bootknife} & \multicolumn{2}{|c|}{GAN} \\
\cline{2-9}
& MSE & MAE & MSE & MAE & MSE & MAE & MSE & MAE \\
 \hline
 Exp/Sinusoidal & 7.2881 & 2.696 & 7.1627  &  2.6726 &  7.1494 &  2.6698 & \textbf{7.1163}  & \textbf{2.6634}  \\
 Exp/Power Law & 3.767 & 1.9375 & \textbf{3.74}  & \textbf{1.9307} &  3.7777 & 1.9396  & 3.9508  & 1.984  \\
 LN/Sinusoidal & 10.078 & 3.1507 & 9.4833  & 3.0623  & 9.7539  &  3.105 & \textbf{8.6918}  & \textbf{2.9343}  \\
 LN/Power Law & 4.7818 & 2.1693 & 4.8746  &  2.187 & 4.762  & 2.1668  &  \textbf{4.6401} & \textbf{2.1434}  \\
 $\Gamma$/Sinusoidal & 7.3101 & 2.6983 & 7.1604  &  2.6708 & 7.2172  & 2.6817  & \textbf{7.1054}  &  \textbf{2.6609} \\
 $\Gamma$/Power Law & 3.8133 & 1.9485 &  \textbf{3.7524} & \textbf{1.9333}  & 3.8247  & 1.9514  &  3.9519 & 1.9839  \\
 Weib/Sinusoidal & 7.1338 & 2.6653 &  7.1083 &  2.6608 & \textbf{7.0985}  & \textbf{2.6592}  & 7.1294  &  2.665 \\
 Weib/Power Law & 3.7214 & 1.9246 & \textbf{3.6852}  &  \textbf{1.9152} & 3.6964  & 1.9188  & 3.9009  &  1.971 \\
 \hline
\end{tabular}
\label{tab_errors}
\end{footnotesize}
\end{table}

And for both resampling methods, i.e. the bootstrap and bootknife, we apply a similar approach (see Algorithm \ref{resalg}).
In this case, $B=500$ bootstrap (or bootknife, respectively) samples were generated using the training data about the value of the claims only.
Obviously, the moments of catastrophes can't be resampled similarly to avoid the impossible situation when two (or more) of the claims occur at the same time.
The means of the MLEs obtained with the bootstrap and bootknife for the probability distributions of $C_i$ can be found in Table~\ref{tab_values_claims}.
As we can see, these estimators are similar to the ones for the no-resampling approach, especially in the case of the bootknife method.

\begin{algorithm}[htb]
\caption{Resampling algorithm}
\label{resalg}
\begin{algorithmic} 
\STATE {Provide the training dataset $\mathbb{C} = \{ C_1, \ldots, C_n\} $}
    \FORALL{$b \in \{1, 2, \dots, B\}$}
        \STATE Generate randomly the bootstrap/bootknife sample $\mathbb{C}^{*b} = \left \{ C_1^{*j}, \ldots, C_n^{*j} \right \} $ based on $\mathbb{C}$
        \STATE Calculate the MLEs  using $\mathbb{C}^{*b}$ for the fixed probability distribution of the value of the claims
    \ENDFOR
\RETURN The mean values of the obtained MLEs
\end{algorithmic}
\end{algorithm}

Then, using the obtained MLEs and parameters for the intensity function that are the same as for the no-resampling method (see Table \ref{tab_int_claims}), the respective errors for the different models of $C_i$ and $N_t$ can be estimated (see the second and third row in Table \ref{tab_errors}).
It seems that the bootstrap slightly improves the fitting of these models to the test data in many cases.

The DATGAN approach (after its training) enables us to generate new samples for both the values and moments of the claims (i.e., $C_i$ and $t_i$ independently, but simultaneously).
As noted in Sect. \ref{gainandres}, these new samples should ``mimic'' the behavior and statistical characteristics of our training set.
Then, based on the generated values, the new estimators for both the models of the claim values and their times of occurrences can be obtained (see Algorithm \ref{ganalg}).
The simulated means can be found in Tables \ref{tab_values_claims} and \ref{tab_int_claims}, respectively.
The calculated MLEs for the probability distributions of $C_i$ differ significantly if they are compared with the outputs of the resampling approaches.
The parameters of the intensity functions $\lambda (t)$ are rather close to their no-resampling counterparts.
Then, as in the previous cases, the errors for the test data can be estimated using simulations (see the last column in Table \ref{tab_errors}).
It seems that for the GAN approach, these errors for many pairs of models are the lowest ones.
To increase the readability, the lowest values of the errors for each model (i.e. each row) are emboldened in Table \ref{tab_errors}.
The estimated probability of the ruin for 5 and 10 years horizon is equal to zero for the above-mentioned assumptions.

\begin{algorithm}[htb]
\caption{GAN algorithm}
\label{ganalg}
\begin{algorithmic} 
\STATE {Provide the training dataset $(\mathbb{C}, \mathbf{t}) = \{ (C_1, t_1), \ldots, (C_n, t_n)\} $}
\STATE {Train the GAN model using $(\mathbb{C}, \mathbf{t})$}
    \FORALL{$b \in \{1, 2, \dots, B\}$}
        \STATE Generate randomly the additional sample $\left (\mathbb{C}^{*b}, \mathbf{t}^*  \right ) = \left \{ \left (C_1^{*j}, t_1^{*j} \right), \ldots,  \left (C_n^{*j}, t_n^{*j} \right )  \right \} $
        \STATE Calculate the MLEs  using $\mathbb{C}^{*b}$ for the fixed probability distribution of the value of the claims
        \STATE Calculate the estimators of the intensity function $\lambda (t)$ using $\mathbf{t}^{*b}$ and \eqref{nhppestim}
    \ENDFOR
\RETURN The mean values of the obtained MLEs and estimators  of the intensity function
\end{algorithmic}
\end{algorithm}

Then, it seems that the GAN method improves the fitting of the models in  most cases, and the bootstrap takes second place in this field.
Overall, the smallest errors are obtained for the Power Law intensity function coupled with the Weibull distribution. 
Slight improvements of the models may be seen as negligible, but we should remember that MSE is measured in {\$}$10^{16}$, and the MAE -- in {\$}$10^{8}$.
Therefore, the gains for the insurer related to a better fit of the model can be even billions of dollars.

\subsection{Fuzzy expert's opinion}
\label{fuexop}

The values of the claim process and errors of the fitted models described in Sect.~\ref{noresap} are based on the respective means and seem to be ``crisp'' and precise results.
However, this is not the case -- because of the embedded randomness of the conducted simulations, the mean is only one (however very important) of many other ``possible answers''.
Therefore, the insurer may be interested in a result that is imprecise but provides more information (in some sense)  instead, e.g., about the possible error.
Then, it may be profitable to apply fuzzy numbers to enrich our obtained solutions and gain, let's say, an ``expert's opinion''.

To replace the ``crisp'' estimated mean of the value of the claims $\hat{S}_T$ with its fuzzy counterpart, we propose an approach related to the construction of a series of the $\alpha$-cuts.
Similar procedures exist in the literature \cite{Romaniuk201733,Romaniuk2019437,Romaniuk2020352}.
As the $\alpha$-cut of the fuzzy output $\tilde{S}_T$ for the given $\alpha$, we use an interval
\begin{equation}
    \tilde{S}_{T,\alpha} =\left  [ q_{\alpha / 2} \left (\hat{S}_{T,1} , \ldots, \hat{S}_{T,i} \right ) , q_{1-\alpha / 2} \left (\hat{S}_{T,1} , \ldots, \hat{S}_{T,i} \right ) \right  ] ,
\end{equation} 
where $q_x$ denotes the quantile of the order $x \in [0,1]$.
Then, the core of the LRFN $\tilde{S}_T$ is given by the median of the simulated values of the claim process, and its support -- by 0{\%} and 100{\%} quantiles (for the left- and right-hand side of the interval, respectively).

\begin{figure}[htb]
\centering
\subfloat[Model: Weib/Power Law \label{compfiga}]{%
  \includegraphics[width=0.48\textwidth]{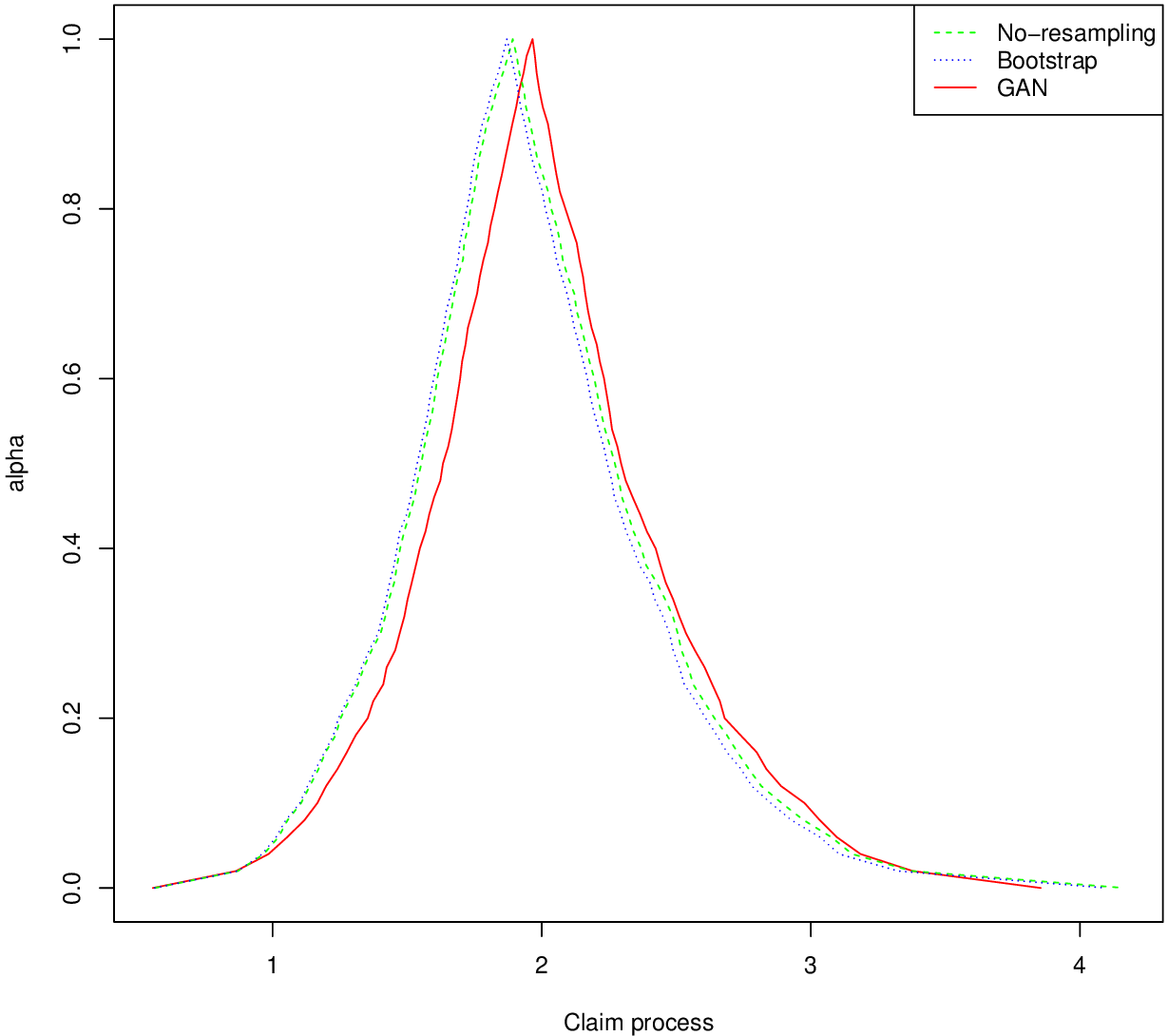}%
}\hfil
\subfloat[Model: LN/Power Law \label{compfigb}]{%
  \includegraphics[width=0.48\textwidth]{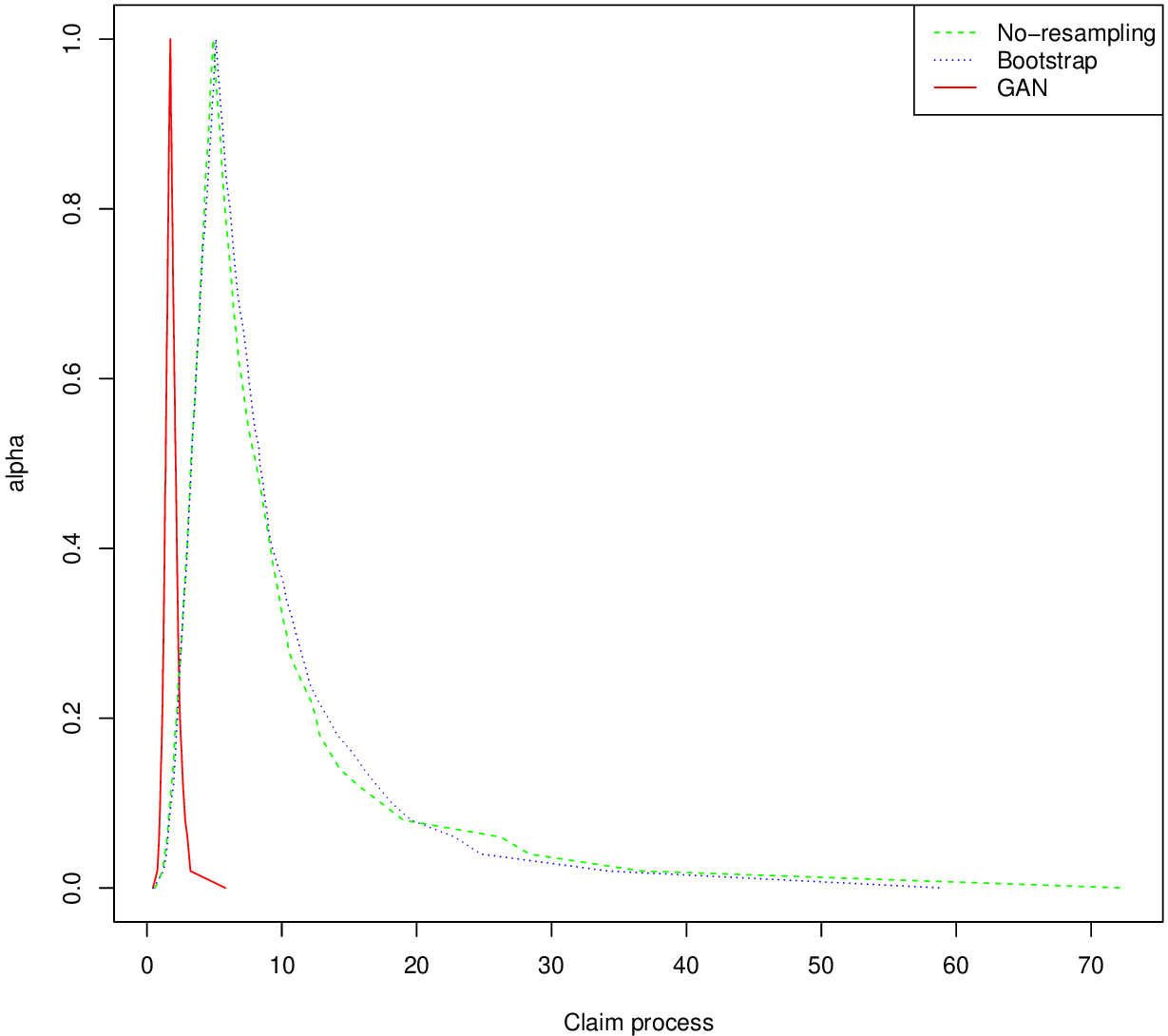}%
}

\caption{Comparisons of fuzzy opinions concerning the value of the claim process.}
\label{compfig}

\end{figure}

The obtained LRFN gives us additional insight into the considered problem (see Fig. \ref{compfig} for some examples measured in \${$10^7$}).
To simplify our reasoning, the one-year time horizon is set (i.e. $T=1$).
In the case of  ``the best-fitted model'' (i.e., the pair: the Weibull distribution of the claim values and the Power Law intensity function), the simulated LRFNs are similar and rather close to one another,  $\tilde{S}_T$ for the GAN method is slightly moved to the right (see Fig. \ref{compfiga}).
If the lognormal distribution is coupled with the Power Law function, the results are quite different (see Fig. \ref{compfigb}).
The LRFN obtained for the GAN model is quite ``thin'' and strongly shifted to the left, while the LRFNs for both the no-resampling and bootstrap approaches are right-skewed and have wide supports so they indicate the high possible error.
Then, it seems that depending on the selected model, the answers can be quite different.
If necessary, the LRFNs can be approximated then by other types of FNs, e.g., triangular or trapezoidal ones~\cite{Grzegorzewski2010}.

\section{Conclusions}
\label{con}

In this paper, we compared three methods of improving the quality of the catastrophic data -- the bootstrap, bootknife, and GAN approaches.
As the numerical experiments suggest, the GAN method gives the best fit for the considered models of the claim values and occurrences.
Additionally, we proposed enriching the estimator of the value of the claim process with its fuzzy counterpart based on the left-right fuzzy numbers.

Further experiments and an introduction of different resampling methods can be valuable.
Moreover, the problem of the imputation of the catastrophic dataset (instead of its only ``reusing'' and ``multiplication'') is still a challenging one.

%
%
%
\bibliographystyle{abbrv}
\bibliography{art_claims_2023_ndmr_bib}

\begin{thebibliography}{10}

\bibitem{doi:10.1142/7431}
S.~Asmussen and H.~Albrecher.
\newblock {\em Ruin Probabilities}.
\newblock World Scientific, 2nd edition, 2010.

\bibitem{ban_coroianu_pg}
A.~Ban, L.~Coroianu, and P.~Grzegorzewski.
\newblock {\em Fuzzy Numbers: Approximations, Ranking and Applications}.
\newblock Polish Academy of Sciences, Warsaw, 2015.

\bibitem{Chernobai2006}
A.~Chernobai, K.~Burnecki, S.~Rachev, S.~Tr{\"u}ck, and R.~Weron.
\newblock Modelling catastrophe claims with left-truncated severity
  distributions.
\newblock {\em Computational Statistics}, 21(3):537--555, Dec 2006.

\bibitem{Efron}
B.~Efron.
\newblock Bootstrap methods: Another look at the jackknife.
\newblock {\em Annals of Statistics}, 7:1--26, 1979.

\bibitem{GIURICICH2019498}
M.~N. Giuricich and K.~Burnecki.
\newblock Modelling of left-truncated heavy-tailed data with application to
  catastrophe bond pricing.
\newblock {\em Physica A: Statistical Mechanics and its Applications},
  525:498--513, 2019.

\bibitem{NIPS2014_5ca3e9b1}
I.~Goodfellow, J.~Pouget-Abadie, M.~Mirza, B.~Xu, D.~Warde-Farley, S.~Ozair,
  A.~Courville, and Y.~Bengio.
\newblock Generative adversarial nets.
\newblock In Z.~Ghahramani, M.~Welling, C.~Cortes, N.~Lawrence, and
  K.~Weinberger, editors, {\em Advances in Neural Information Processing
  Systems}, volume~27, pages 2672--2680. Curran Associates, Inc., 2014.

\bibitem{Grzegorzewski2010}
P.~Grzegorzewski.
\newblock Algorithms for trapezoidal approximations of fuzzy numbers preserving
  the expected interval.
\newblock In B.~Bouchon-Meunier, L.~Magdalena, M.~Ojeda-Aciego, J.-L. Verdegay,
  and R.~R. Yager, editors, {\em Foundations of Reasoning under Uncertainty},
  pages 85--98, Berlin, Heidelberg, 2010. Springer Berlin Heidelberg.

\bibitem{Grzegorzewski20201650}
P.~Grzegorzewski, O.~Hryniewicz, and M.~Romaniuk.
\newblock Flexible bootstrap for fuzzy data based on the canonical
  representation.
\newblock {\em International Journal of Computational Intelligence Systems},
  13(1):1650--1662, 2020.

\bibitem{Grzegorzewski2022285}
P.~Grzegorzewski and M.~Romaniuk.
\newblock Bootstrap methods for epistemic fuzzy data.
\newblock {\em International Journal of Applied Mathematics and Computer
  Science}, 32(2):285--297, 2022.

\bibitem{Hesterberg2002UnbiasingTB}
T.~Hesterberg.
\newblock Unbiasing the bootstrap - bootknife sampling vs. smoothing.
\newblock 2002.

\bibitem{nhess-12-535-2012}
W.~Kron, M.~Steuer, P.~L\"ow, and A.~Wirtz.
\newblock How to deal properly with a natural catastrophe database - analysis
  of flood losses.
\newblock {\em Natural Hazards and Earth System Sciences}, 12(3):535--550,
  2012.

\bibitem{lederrey2022datgan}
G.~Lederrey, T.~Hillel, and M.~Bierlaire.
\newblock {DATGAN}: Integrating expert knowledge into deep learning for
  synthetic tabular data, 2022.

\bibitem{Romaniuk201733}
M.~Romaniuk.
\newblock Analysis of the insurance portfolio with an embedded catastrophe bond
  in a case of uncertain parameter of the insurer's share.
\newblock {\em Advances in Intelligent Systems and Computing}, 524:33--43,
  2017.

\bibitem{Romaniuk2019437}
M.~Romaniuk.
\newblock On some applications of simulations in estimation of maintenance
  costs and in statistical tests for fuzzy settings.
\newblock In A.~Steland, E.~Rafaj{\l}owicz, and O.~Okhrin, editors, {\em
  Stochastic Models, Statistics and Their Applications}, pages 437--448, Cham,
  2019. Springer International Publishing.

\bibitem{Romaniuk2019111}
M.~Romaniuk.
\newblock Simulation-based analysis of penalty function for insurance portfolio
  with embedded catastrophe bond in crisp and imprecise setups.
\newblock {\em Advances in Intelligent Systems and Computing}, 854:111--121,
  2019.

\bibitem{Romaniuk20203}
M.~Romaniuk.
\newblock Imprecise approaches to analysis of insurance portfolio with
  catastrophe bond.
\newblock {\em Communications in Computer and Information Science}, 1239
  CCIS:3--16, 2020.

\bibitem{RJ-2023-036}
M.~Romaniuk and P.~Grzegorzewski.
\newblock Resampling fuzzy numbers with statistical applications:
  {F}uzzy{R}esampling package.
\newblock {\em The {R} Journal}, 15:271--283, 2023.

\bibitem{Romaniuk2020352}
M.~Romaniuk and O.~Hryniewicz.
\newblock Estimation of maintenance costs of a pipeline for a {U}-shaped hazard
  rate function in the imprecise setting.
\newblock {\em Eksploatacja i Niezawodnosc}, 22(2):352--362, 2020.

\bibitem{DEMCart}
K.~M. Rychlik and M.~Romaniuk.
\newblock Improved {DE}-{MC} algorithm with automated outliers detection.
\newblock In S.~Massanet, S.~Montes, D.~Ruiz-Aguilera, and
  M.~Gonz{\'a}lez-Hidalgo, editors, {\em Fuzzy Logic and Technology, and
  Aggregation Operators}, pages 701--712, Cham, 2023. Springer Nature
  Switzerland.

\bibitem{GANart}
Z.~Wang, Q.~She, and T.~E. Ward.
\newblock Generative adversarial networks in computer vision: A survey and
  taxonomy.
\newblock {\em ACM Computing Surveys}, 54(2), 2021.
\newblock Article 37.

\bibitem{5451114}
B.~M. Wilamowski and H.~Yu.
\newblock Improved computation for {L}evenberg-{M}arquardt training.
\newblock {\em IEEE Transactions on Neural Networks}, 21(6):930--937, 2010.

\bibitem{PowerLaw}
T.-H. Yoo.
\newblock The infinite {NHPP} software reliability model based on monotonic
  intensity function.
\newblock {\em Indian Journal of Science and Technology}, 8:1--7, 2015.

\end{thebibliography}

\end{document}